\documentclass{article}

\pdfoutput=1

\usepackage[preprint]{neurips_2023}

\usepackage[toc,page]{appendix}
\usepackage[utf8]{inputenc} 
\usepackage[T1]{fontenc}    
\usepackage{hyperref}       
\usepackage{url}            
\usepackage{booktabs}       
\usepackage{amsfonts}       
\usepackage{nicefrac}       
\usepackage{microtype}      
\usepackage{xcolor}         
\usepackage{graphicx} 
\usepackage{amsmath}
\usepackage{makecell}
\usepackage{mathtools}
\usepackage{multirow}
\usepackage{hhline}
\usepackage{cellspace}
\title{Unsupervised Coordinate-Based Video Denoising}

\author{%
  Mary Damilola Aiyetigbo\thanks{Clemson University, School of Computing} \\
  \texttt{maiyeti@clemson.edu} \\
  \And
  Dineshchandar Ravichandran\footnotemark[1] \\
  \texttt{dravich@clemson.edu} \\
  \AND
  Reda Chalhoub\thanks{MUSC, Department of Neuroscience} \\
  \texttt{chalhoub@musc.edu} \\
  \And
  Peter Kalivas\footnotemark[2] \\
  \texttt{kalivasp@musc.edu} \\
  \And
  Nianyi Li\footnotemark[1] \\
  \texttt{nianyil@clemson.edu} \\
}

\begin{document}

\maketitle

\begin{abstract}

In this paper, we introduce a novel unsupervised video denoising deep learning approach that can help to mitigate data scarcity issues and shows robustness against different noise patterns, enhancing its broad applicability. Our method comprises three modules: a Feature generator creating features maps, a Denoise-Net generating denoised but slightly blurry reference frames, and a Refine-Net re-introducing high-frequency details. By leveraging the coordinate-based network, we can greatly simplify the network structure while preserving high-frequency details in the denoised video frames. 
Extensive experiments on both simulated and real-captured demonstrate that our method can effectively denoise real-world calcium imaging video sequences without prior knowledge of noise models and data augmentation during training.

\end{abstract}

\section{Introduction}
Video denoising is a critical task in computer vision, with various applications such as video surveillance, medical imaging, and autonomous driving. The goal is to remove the noise present in videos, which can be caused by various factors such as sensor noise, compression, and low-light conditions. Supervised Convolutional Neural Networks (CNNs) have been successful in solving this problem by learning the mapping between noisy and clean images from large datasets \cite{dvdnet,fastdvdnet,rvidenet,deformableattention}. However, obtaining such datasets can be challenging and time-consuming, especially when the noise is complex and diverse.

To address this challenge, unsupervised video denoising methods have emerged in recent years, which do not require labeled data for training. One such approach is based on the "blind-spot" technique, where the CNN is trained to estimate each noisy pixel from its surrounding spatial neighborhood without considering the pixel itself \cite{udvd}. However, most of these self-supervised methods require either training on synthetic datasets by adding random noise to the clean image during training, or need long video sequences to learn the implicit noise model \cite{udvd,noise2void}. 
In applications like microscopy, where video sequences might be short, obtaining noiseless ground truth videos and estimating the underlying tractable model of the noise pose significant challenges.
In such cases, the performance of denoising methods based on "blind-spot" CNN can significantly degrade \cite{self-supervised}. Therefore, developing an unsupervised video denoising method that can work directly on raw noisy videos is critical to improve the accuracy and robustness of video denoising in challenging scenarios.

\begin{figure*}
  \centering
  \includegraphics[width=1\linewidth]{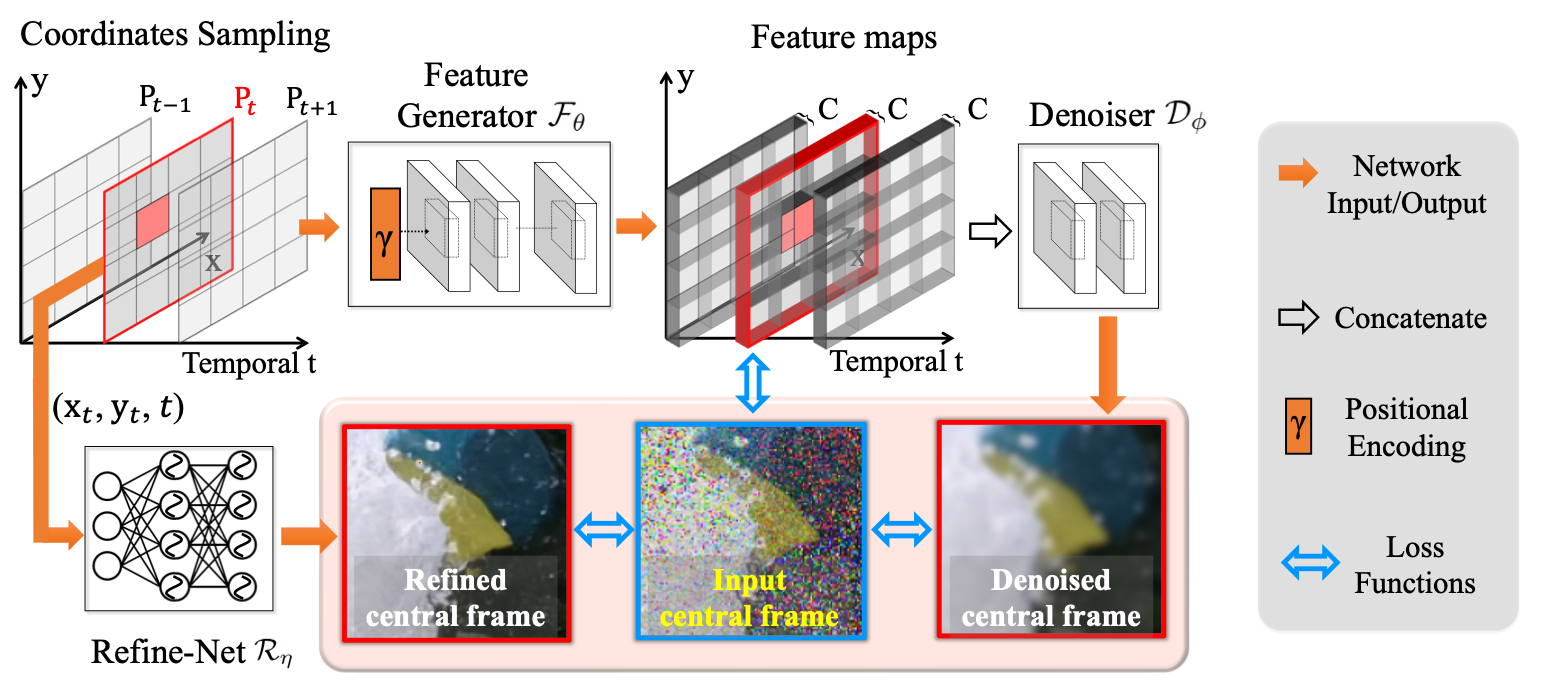}
  \caption{The pipeline consists of three main components: a feature generator $\mathcal{F}$, which renders feature maps matching the input frame coordinates; a Denoise-Net $\mathcal{D}$, which utilizes these feature maps to generate denoised yet slightly blurry reference frames; and a Refine-Net $\mathcal{R}$, which reintroduces high-frequency details to enhance the clarity of the denoised frames. The entire network is trained by optimizing the difference between the generated denoised frame, the input noisy frame, and the final refined output frame, ensuring efficient and accurate video denoising.}
  \label{fig:pipeline}
\end{figure*}

The architecture of our network is shown in Fig.~\ref{fig:pipeline}, comprises three key components, each designed to effectively denoise video data. First, the feature generator $\mathcal{F}$ renders feature maps that align with the coordinates of the input frames. Second, the Denoise-Net $\mathcal{D}$ utilizes these feature maps to produce denoised yet slightly blurry reference frames. Finally, the Refine-Net $\mathcal{R}$ restores high-frequency details to the denoised frames, enhancing overall image clarity. The network is trained by minimizing the discrepancy between the generated denoised frame, the input noisy frame, and the refined final output frame, ensuring both efficient denoising and preservation of the original video data's integrity and quality. To streamline the network architecture and enhance training efficiency, we incorporate coordinate-based networks, as referenced in \cite{chibane2020implicit,dupont2020equivariant,riegler2020free,fourier,siren}, into both $\mathcal{D}$ and $\mathcal{R}$. We conduct comprehensive experiments on a diverse range of noisy videos, including both simulated and real-captured footage. In comparison to state-of-the-art method, our approach demonstrates superior performance in effectively correcting noise, highlighting its efficacy and potential for widespread application.


\section{Related Works}
\noindent \textbf{Supervised Video Denoising.} 
Supervised CNN approaches to denoising, such as those presented in \cite{dncnn, ffdnet, videnn, rvidenet}, have achieved state-of-the-art results in both single image and video denoising, requiring clean images for model training. Several techniques, like residual learning \cite{residual} and batch normalization \cite{batchnorm} in DnCNN \cite{dncnn}, or the use of dilated convolution \cite{dilated} for a faster model, have been developed. The introduction of downsampling and upsampling layers in FFDNet \cite{ffdnet} aimed to increase the receptive field. Two-stage denoising without explicit motion estimation was explored by DVDNet \cite{dvdnet}, ViDeNN \cite{videnn}, and FastDVDnet \cite{fastdvdnet}. However, these models' reliance on unrealistic noisy/clean pairs is a drawback, especially in medical imaging where clean ground truth images are seldom available.

\noindent \textbf{Unsupervised Video Denoising.} 
Unsupervised denoising algorithms like Noise2Noise (N2N) \cite{noise2noise}, Frame2Frame (F2F) \cite{frame2frame}, and Multi Frame2Frame (MF2F) \cite{mf2f} leverage noisy images as both inputs and targets, or consecutive frames as noisy targets for denoising. Techniques like the blind spot technique in Noise2Void (N2V) \cite{noise2void} and the convolution blind-spot network architecture \cite{self-supervised} have been developed to estimate the underlying clean signal. While UDVD \cite{udvd} achieved state-of-the-art results, their methodology necessitates noise addition at every model iteration and substantial data augmentation. Furthermore, the deep interpolation algorithm by Lecoq et al. \cite{deepinter} required over 200,000 data samples and showed limitations in generalizability.
In contrast to these methods, our approach can effectively denoise short video sequences and generalize to different types of noisy videos without requiring excessive data augmentation or iterative noise addition.


\noindent \textbf{Implicit Neural Representation.} 
Implicit neural representations, also known as coordinate-based representations, utilize fully connected neural networks to associate input coordinates with their corresponding signal values. They have shown remarkable utility across a range of tasks, including view synthesis \cite{nerf}, image representation \cite{fourier, siren}, and 3D shape representation \cite{neural}. One significant development in this domain is the Sinusoidal Representation Networks (SIREN) \cite{siren}, which leverage periodic activation functions to encode positional information and model complex natural signals with high precision. This approach allows for rapid training convergence, which significantly improves efficiency, especially in resource-intensive applications. In our framework, we employ coordinate-based networks in the feature generator $\mathcal{F}$ and Refine-Net $\mathcal{R}$ to enhance the denoising performance. 

\section{Unsupervised Coordinate-Based Video Denoising}
\label{sec:method}

Given a sequence of noisy video frames $\{I_t | t = 1, 2, \dots, N\}$ and their corresponding coordinates $\{G_t | t = 1, 2, \dots, N\}$, where $G_t(\textbf{p}_t )$ represents the coordinates of pixel $\textbf{p}_t = (x, y,t)$ in the noisy frame $I_t$, our objective is to recover the noise-free video frames $\{J_t | t = 1, 2, \dots, N\}$. Our approach consists of three primary components: a feature generator $\mathcal{F}_\theta$, a Denoise-Net $\mathcal{D}_\phi$, and a Refine-Net $\mathcal{R}_\eta$, as illustrated in Fig.~\ref{fig:pipeline}.

\subsection{Feature Generator $\mathcal{F}_\theta$}
Our Feature Generator processes a batch of uniformly sampled coordinate grids $\{G_t | t = 1,...,B\}$, where each $G_t \in\mathbb{R}^{H \times W \times 3}$, to generate a corresponding batch of feature maps $\{F_t | t = 1,...,B\}$, each being $F_t \in\mathbb{R}^{H \times W \times C}$. Here, $B$ is the batch size, while $C$ is the number of feature channels.

Before the coordinate grid is fed into $\mathcal{F}_\theta$, each coordinate $\textbf{p}_t=(x,y,t)$ is subject to positional encoding \cite{nerf,mipnerf, rawnerf}. This encoding step transforms low-dimensional input coordinates into a higher-dimensional space, enabling the model to better learn and represent high-frequency details inherent in the image data. The positional encoding function we adopt is given as:

\begin{equation}
\gamma(\mathbf{p}_t) = [\mathrm{sin}(2^0 \pi \mathbf{p}_t), \mathrm{cos}(2^0 \pi \mathbf{p}_t), ..., \mathrm{sin}(2^{L-1} \pi \mathbf{p}_t), \mathrm{cos}(2^{L-1} \pi \mathbf{p}_t)]
\label{posenc}
\end{equation}

In this equation, $L$ serves as a hyperparameter controlling the level of detail or high-frequency information in the output. By selecting a smaller value for $L$, we can effectively reduce the level of high-frequency noise in the image data, as noise often manifests as high-frequency information. For our experiments, we set $L=30$. The input coordinates, normalized to the range $[-1, 1]$ using a mesh grid, are passed through the encoding function $\gamma(.)$. The resulting high-dimensional output $\gamma(G_t)\in\mathbb{R}^{H \times W \times 6L}$ is subsequently fed into the feature generator $\mathcal{F}_\theta$:
\begin{equation}
    F_t = \mathcal{F}_\theta(\gamma(G_t))
\end{equation}
where $F_t\in\mathbb{R}^{H \times W \times C}$ is the feature maps corresponding to each noisy frame, and $C$ denotes the channel size of the output features.

Our feature generator is composed of 6 convolution layers, each featuring a kernel size of 3 and identical padding to maintain the size throughout the model layers. Each layer has 256 feature channels with batch normalization (BN) applied solely to the first two layers. ReLU activation is employed for all layers, except for the last one.
\subsection{Denoiser $\mathcal{D}_\phi$}

The denoiser network takes the concatenated feature maps output from the feature generator as input, and generates a denoised central frame $\hat{I}_B$:
\begin{equation}
    \hat{I}_B = \mathcal{D}_\phi([F_1,F_2,...F_B])
\end{equation}
where the concatenation is applied along the feature channel dimension. This allows $\mathcal{D}_\phi$ to learn spatial-temporal patterns along the neighboring time frames eliminate the noise. Note that the output of $\mathcal{D}_\phi$ may be somewhat blurred, because we've set a low $L$ in the feature generator. 

The architecture of $\mathcal{D}_\phi$ includes 6 convolutional layers. ReLU activation is used in the first five layers, and a sigmoid activation function is used in the last layer. Unlike in the feature generator, batch normalization is not applied in this stage. Each layer uses 256 filters with a kernel size of 3, except for the last two layers which have 96 and the number of color channels filters, respectively, with kernel sizes of 1.


\subsection{Refine-Net $\mathcal{R}_\eta$}
The refine-net is built upon the backbone of the Sinusoidal Representation Networks (SIREN) \cite{siren}, which is a type of coordinate-based network that uses periodic activation functions, particularly sine functions, in place of traditional activation functions like ReLU. SIREN's unique characteristic lies in its ability to naturally model the high-frequency details of complex patterns by leveraging its intrinsic periodic activation functions. In our context, $\mathcal{R}_\eta$ uses the SIREN network to take the coordinates grid of the central frame as input and generate the refined image $\hat{I}_R$. 

\begin{equation}
    \hat{I}_R = \mathcal{R}_\eta(G_t^c)
\end{equation}
where $G_t^c$ is the coordinates of the central frames in the input batch. 

Note that the SIREN-based refine-net is particularly beneficial in our case, as it helps to further refine the denoised output from $\mathcal{D}_\phi$ by enhancing the finer details and correcting any blurring introduced in the denoising process. The output from $\mathcal{R}_\eta$ represents the final denoised and refined video frames.

\section{Network Optimization}

Given the unsupervised nature of our architecture, the network optimization problem is highly non-convex with a vast parameter search space. To navigate this challenge, we propose a two-step network optimization strategy that exploits the structural similarity between neighboring frames to reconstruct the central frame.

\begin{figure}[t!]
  \centering
  \includegraphics[width=1\textwidth]{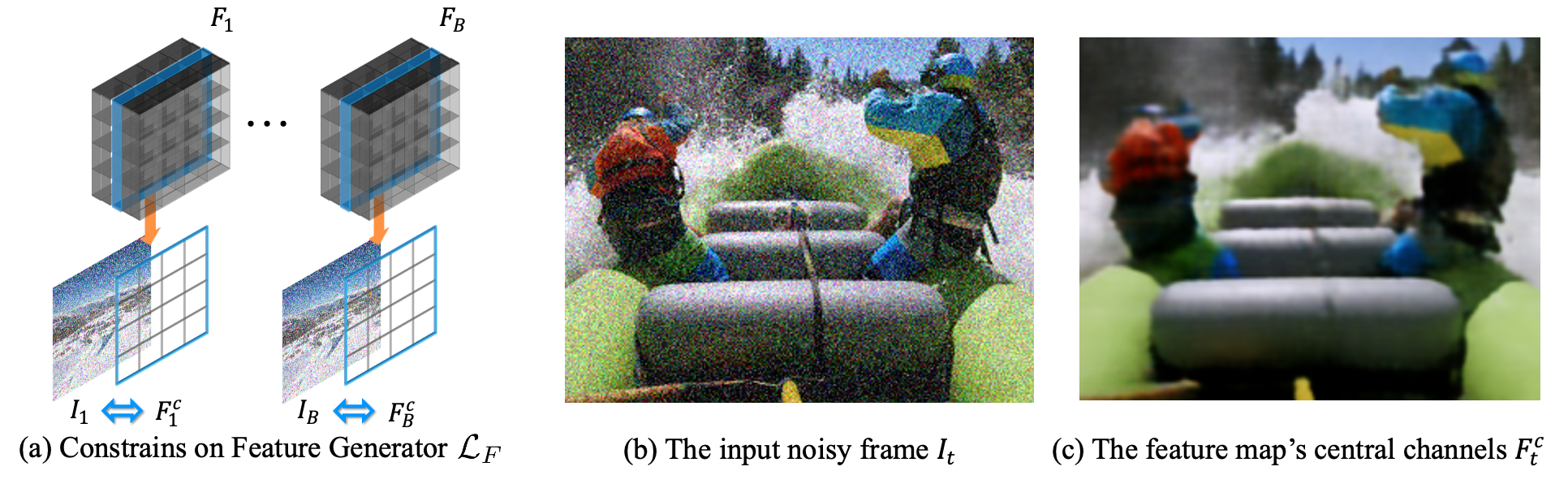}
  \caption{The illustration of $\mathcal{L}_F$}
  \label{fig:illu}
\end{figure}

\subsection{First Stage: Joint Training of the Feature Generator $\mathcal{F}_\theta$ and Denoiser $\mathcal{D}_\phi$}
In the first stage, we jointly train the feature generator $\mathcal{F}_\theta$ and the denoiser $\mathcal{D}_\phi$ in an end-to-end fashion. The objective function for this stage is composed of two parts. The first part is the $l_1$ loss, which measures the difference between the denoised central frame $\hat{I}_B$ and the central frame $I^c$ of the input batch of noisy frames $\{I_t | t = 1, 2, \dots, B\}$:

\begin{equation}
\mathcal{L}_{\mathcal{D}} = ||\hat{I}_B - I^c||
\end{equation}

The second part of the objective function ensures that the feature maps generated by $\mathcal{F}_\theta$ capture the image information. This is achieved by enforcing similarity between the central channels of each feature map and the corresponding noisy frame:

\begin{equation}
\mathcal{L}_{\mathcal{F}} = \frac{1}{B}\sum_{t=1}^B|| F_t^{c} - I_t||
\end{equation}

where $F_t^{c}$ is the central channels of each feature map $\mathcal{F}_\theta(\gamma(G_t))$  and $I_t$ is the corresponding noisy frame, as shown in Fig.~\ref{fig:illu} (a).

The final loss function for the first stage is the sum of these two losses:

\begin{equation}
\mathcal{L}_{1} = \mathcal{L}_{\mathcal{D}} + \lambda_1 \mathcal{L}_{\mathcal{F}}
\end{equation}

where $\lambda_1$ is a weight parameter that controls the trade-off between the two terms. To ensure our model doesn't simply learn to reproduce the noise present in the input, we train the network for approximately 2000 epochs.

\subsection{Second Stage: Training the Refine-Net $\mathcal{R}_\eta$}
In the second stage, we focus on training the Refine-Net $\mathcal{R}_\eta$. Unlike in the first stage, where $\mathcal{F}_\theta$ and $\mathcal{D}_\phi$ were trained together, here we fix $\mathcal{F}_\theta$ and $\mathcal{D}_\phi$ and solely train $\mathcal{R}_\eta$.

We use the coordinates of the central frame $G_t^c$ as input to $\mathcal{R}_\eta$, which outputs a refined denoised image $\hat{I}_R$. As this network is designed to further improve the quality of the denoised frames, the loss function for this stage is defined to measure the difference between the output of $\mathcal{R}_\eta$ and both the noisy central frame $I_t^c$ and the denoised central frame $\hat{I}_B$ from the first stage. Specifically, the loss function is defined as:

\begin{equation}
L_{2} = \lambda_2||\mathcal{R}\eta(G_t^c) - I^c|| + \lambda_3||\mathcal{R}\eta(G_t^c) - \hat{I}_B||,
\label{siren_loss}
\end{equation}

where $\lambda_2$ and $\lambda_3$ are weight parameters controlling the contribution of each term. These parameters help balance the network's objectives of reducing noise (by making the output similar to $I_B^c$) and preserving details (by making the output similar to $I_t^c$).

This strategy allows the Refine-Net to leverage the advantages of both the noisy and denoised frames, by enhancing details and suppressing noise. The training of $\mathcal{R}\eta$ is performed until satisfactory results are obtained, typically for about 2000 epochs. It's important to note that the second stage training does not affect the training of $\mathcal{F}_\theta$ and $\mathcal{D}_\phi$, which is crucial for preserving the generalization capability of the whole framework.

Our two-stage optimization can effectively remove the noise in the video frames. The first image on the left represents the initial state, which is typically a noisy video frame. The joint training of the feature generator $\mathcal{F}_\theta$ and the denoiser $\mathcal{D}_\phi$ shows a noticeable reduction in noise (middle frame), but some blurriness might still be present due to the low $L$ we used in our feature generator. The right image showcases the result of the refining stage $\mathcal{R}_\eta$. At this stage, the high-frequency details that might have been lost in the denoising process are recovered, resulting in a crisp, clean frame that retains the original structure and details of the scene, demonstrating the step-by-step improvement of our method.

\section{Experiments}
In this section, we evaluate the performance of our proposed method through extensive experiments. Our method is tested on a variety of video sequences with different types of noise, and the results are compared with state-of-the-art denoising methods to demonstrate its effectiveness.

\subsection{Datasets and Setup}

 Our approach is tested on a variety of datasets, encompassing both synthetic and real-world scenarios, to provide a comprehensive evaluation of its performance. Synthetic data are derived from established benchmarks and intentionally corrupted with diverse types of noise, while real-world data are sourced from calcium imaging experiments. Moreover, we outline the specifics of our computational setup and training parameters, detailing the choices that were made to optimize our model's performance. This thorough experimental setup is aimed at providing a robust assessment of our proposed video denoising method and its potential applicability to different types of video data.
 
\paragraph{Synthetic.}
We employed a variety of benchmark datasets, including the DAVIS dataset \cite{davis}, and SET8\cite{dewil2021self} videos s captured with GoPro camera, to provide a comprehensive evaluation of our algorithm. These datasets include a diverse array of video sequences, each with unique content and characteristics, thus allowing us to test our method under numerous conditions. To challenge our algorithm's robustness, we deliberately introduced various types of noise to the clean video sequences. Gaussian noise was added as per the formula:
\begin{equation}
    N(\textbf{c}|\mu,\sigma^2) = \frac{1}{\sqrt{2\pi\sigma^2}}e^{-\frac{(\textbf{c}-\mu)^2}{2\sigma^2}}
\end{equation}
where $\textbf{c}$ is a pixel value, and $\mu$ and $\sigma^2$ are the mean and variance of the Gaussian distribution, respectively \cite{laine2019high}. In our experiment, we set $\mu=0$. Poisson noise was added according to the equation:
\begin{equation}
    P(\textbf{c}|\lambda) = \frac{\lambda^\textbf{c} e^{-\lambda}}{\textbf{c}!},
\end{equation}
where $\lambda$ is the expected number of occurrences \cite{laine2019high}. Salt-and-pepper noise, also known as impulse noise, was added by randomly selecting pixels and setting their values to the minimum and maximum values representing 'salt' and 'pepper' respectively \cite{laine2019high}.

\paragraph{Real-World.}
In addition to the synthetic datasets, we also applied our algorithm to real-world, highly noisy calcium imaging data, as shown in Fig.~\ref{fig:real}. These were locally sourced recordings from freely behaving transgenic mice engaged in cocaine/sucrose self-administration experiments. The recordings were captured using single-channel epifluorescent miniscopes and were subsequently processed using a motion correction algorithm to adjust for translational motion artifacts. This dataset represents the practical complexity and noise levels that are often present in real-world scenarios, further challenging our algorithm's ability to effectively denoise video sequences.

Our proposed method was implemented using the PyTorch framework and trained on an NVIDIA A100 GPU. We employed the Adam optimizer during the training process, with an initial learning rate of $1e-4$ set for the first stage and $1e-5$ for the second stage. Both learning rates were reduced by a factor of 10 every 100 epochs. In our loss function, we set $\lambda_1=0.1$ and $\lambda_2=1.0$ as the balancing factors for our dual-term loss.

\subsection{Quantitative Evaluation}


\begin{figure}[t!]
  \centering
  \includegraphics[width=1\textwidth]{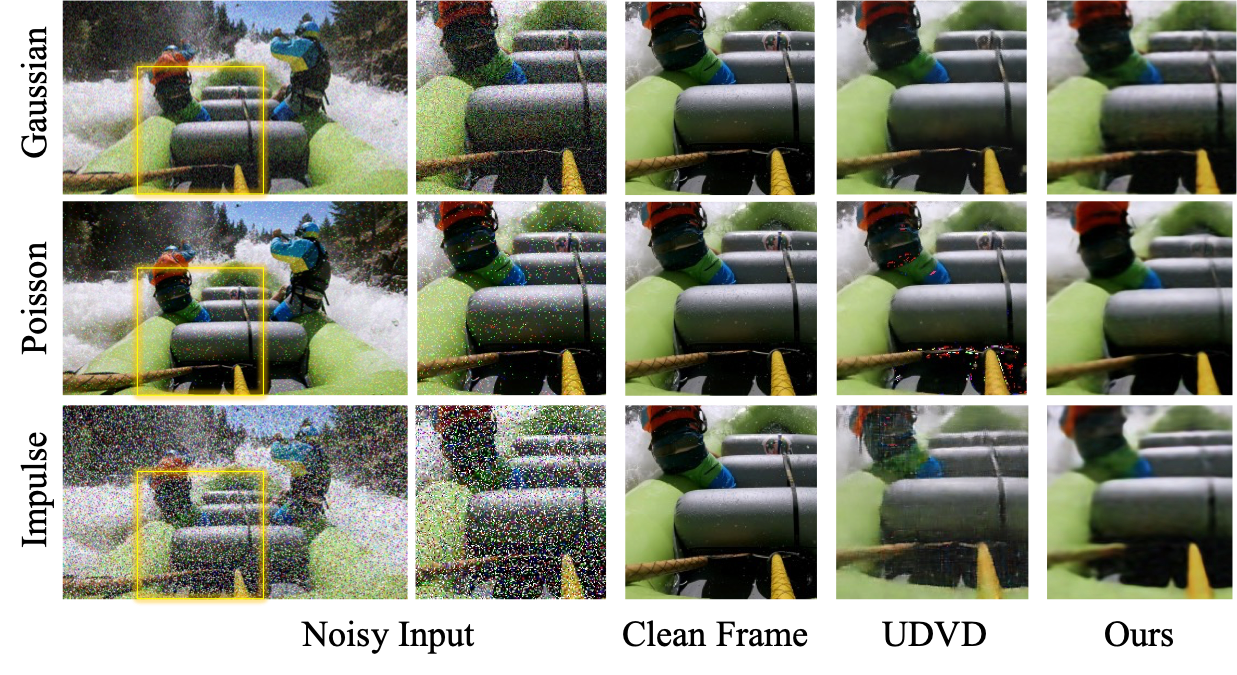}
  \caption{\textbf{Visual comparison of denoising results on the DAVIS dataset:} The figure illustrates denoising results from three types of noise: Gaussian, Poisson, and Impulse. Each row presents the denoised results from our method (right) in comparison with UDVD (middle), along with the original noisy frame (left). It can be observed that our method achieves superior noise removal while preserving intricate details, even in the presence of diverse noise types, outperforming UDVD\cite{udvd}.}
  \label{fig:syn}
\end{figure}

To facilitate a quantitative evaluation of our approach, we employ two widely-accepted metrics in the image and video processing community: the Peak Signal-to-Noise Ratio (PSNR) and Structural Similarity Index Measure (SSIM). These measures allow us to objectively compare our results with those of the current state-of-the-art unsupervised video denoising algorithm, UDVD\cite{udvd}.

The outcomes, as displayed in Table \ref{tab:results}, highlight that our method consistently outperforms UDVD across all tested datasets and noise types. This superior performance is primarily attributed to our approach's effective utilization of spatial-temporal information from the video sequence. Our algorithm leverages this information to efficiently eliminate noise while simultaneously preserving high-frequency details within the frames.

When examining the visual comparison results, as shown in Fig.~\ref{fig:syn}, it becomes clear that UDVD often struggles to effectively remove noise when dealing with short input sequences. This challenge is particularly evident when confronted with Poisson and Impulse noise types, where UDVD tends to produce noticeable artifacts. Conversely, our method shows remarkable resilience even in these demanding situations. We tested our approach on ten short video sequences, each consisting of ten frames, and the results consistently demonstrated our method's superior noise removal capability and robustness.

Overall, the quantitative and qualitative evaluations underscore the potential of our proposed video denoising method as a robust and effective solution, particularly for applications dealing with short, noisy video sequences.

\newcommand{\SH}{\multicolumn{1}{c}{PSNR} & \multicolumn{1}{c}{SSIM}} 
\begin{table}[htbp]
\label{tab:results}
\small
  \centering
  \setlength{\abovetopsep}{12pt}
  \renewcommand{\arraystretch}{1.2}
  \begin{tabular}{*{6}{l}}
    \toprule
    & & \multicolumn{2}{c}{DAVIS} & \multicolumn{2}{c}{SET8}  \\
    \cmidrule(lr){3-4}
    \cmidrule(lr){5-6}
& & \SH & \SH   \\
    \midrule
    \multirow{2}{*}{\parbox{1.2cm}{Gaussian $\sigma=30$}} & UDVD: & 27.9 & \textbf{0.8} & 27.40 & 0.79 \\
    & Ours & \textbf{28.49} & 0.78 & \textbf{29.01} & \textbf{0.80}  \\
    \midrule
    \multirow{2}{*}{\parbox{1.2cm}{Gaussian $\sigma=50$}} & UDVD: & 24.3 & 0.63 & 25.27 & 0.73 \\
    & Ours & \textbf{28.86} & \textbf{0.74} & \textbf{27.36} & \textbf{0.76} \\
    \midrule
    \multirow{2}{*}{\parbox{1.2cm}{Poisson $\lambda=30$}} & UDVD: & 27.7 & \textbf{0.82} & 27.84 & \textbf{0.87} \\
    & Ours & \textbf{29.81} & \textbf{0.82} & \textbf{29.45} & 0.81   \\
    \midrule
    \multirow{2}{*}{\parbox{1.2cm}{Poisson $\lambda=50$}} & UDVD: & 24.3 & 0.77 & 26.30 & \textbf{0.85} \\
    & Ours & \textbf{27.44} & \textbf{0.79} & \textbf{29.05} & 0.81 \\
    \midrule
    \multirow{2}{*}{\parbox{1.2cm}{Impulse $\alpha=0.2$}} & UDVD: & 19.1 & 0.23 & 22.06 & 0.67 \\
    & Ours & \textbf{22.61} & \textbf{0.66} & \textbf{28.57} & \textbf{0.81} \\
    \midrule
    \multirow{2}{*}{\parbox{1.2cm}{Impulse $\alpha=0.3$}} & UDVD: & 16.9 & 0.14 & 19.02 & 0.49 \\
    & Ours & \textbf{20.66} & \textbf{0.62} & \textbf{28.51} & \textbf{0.78} \\
    \bottomrule
  \end{tabular}
  \vspace{5pt}
\caption{\textbf{Denoising results on synthetic noise.} This table presents the PSNR and SSIM values of our model and UDVD on the DAVIS and SET8 video datasets. While UDVD may achieve higher SSIM score in some cases, it does not effectively eliminate the noise, resulting in a generally lower PSNR value. This underscores the superior noise removal capacity of our model, which consistently delivers higher PSNR values across different noise types and datasets.}
\end{table}

\subsection{Qualitative Evaluation}

\begin{figure}[t!]
  \centering
  \includegraphics[width=0.8\textwidth]{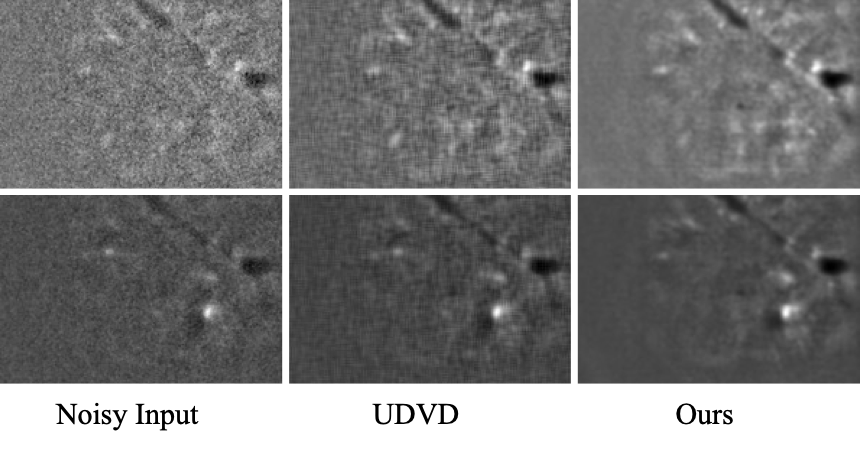}
  \caption{\textbf{Visual comparison on calcium imaging sequences}: The left column shows the original noisy frames, the middle column represents the denoising results from UDVD, and the right column presents the results from our method. It is evident from the comparison that our method is superior at noise reduction, while also maintaining the detailed cellular structures, thereby enhancing the visualization and identification of individual cells.}
  \label{fig:real}
\end{figure}

A visual comparison between our method and UDVD on the highly noisy calcium imaging sequences further underscores our superior performance, as shown in Fig.~\ref{fig:real}. In the noisy frames, it can be challenging to distinguish individual cells due to high levels of noise. UDVD, while effective in reducing some noise, often blurs the intricate cellular structures, making it difficult to identify individual cells.

In contrast, our approach not only removes the noise effectively but also preserves the intricate cellular structures, allowing for better visualization and identification of individual cells. This difference is particularly notable in regions with a high density of cells, where our method is able to maintain the distinct boundaries between cells, whereas UDVD tends to blur them together. This visual comparison highlights our method's ability to handle real-world data with significant noise, offering promising potential for applications in biological and medical imaging.

\subsection{Ablation Study}



\begin{figure}[t!]
  \centering
  \includegraphics[width=1\textwidth]{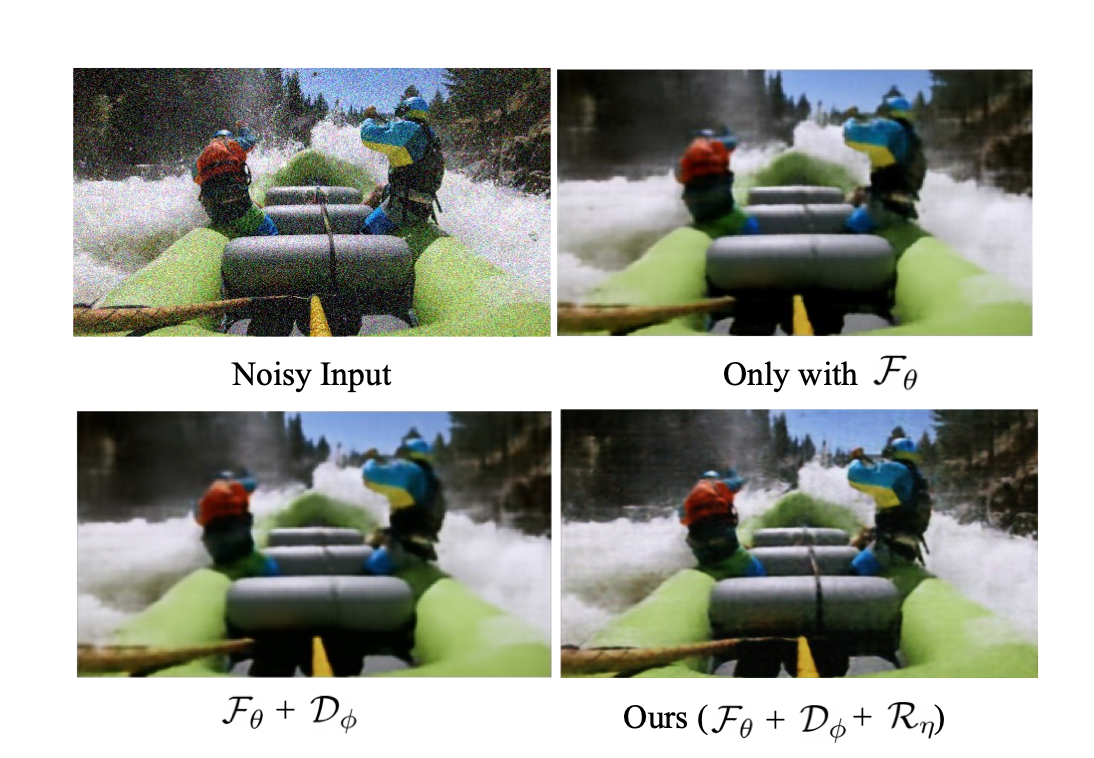}
  \caption{\textbf{Ablation study visual results.}}
  \label{fig:ablation}
\end{figure}

We conduct an ablation study to understand the contribution of each component in our method. Notably, when we omit the refining stage $\mathcal{R}_\eta$, the denoised frames tend to be slightly blurry due to the low $L$ in the feature generator $\mathcal{F}_\theta$. However, with the incorporation of the refining stage, our method is able to effectively recover high-frequency details, thereby underscoring its crucial role in enhancing the overall quality of the denoised frames. We show the visual comparison between each of network variants in Fig.~\ref{fig:ablation}. A detailed presentation of the ablation study results is provided in Table \ref{tab:ablation}.

\begin{table}[h]
    \small
    \begin{center}
    \begin{tabular}{|c|c|c|}
     \hline
     \centering
      Method &  PSNR  & SSIM\\ 
      \hline
      $\mathcal{F}_\theta + \mathcal{D}_\phi$ &  30.50  & 0.81 \\ 
      $\mathcal{F}_\theta + \mathcal{D}_\phi + \mathcal{R}_\eta$  &  \textbf{30.90} & \textbf{0.84} \\
     \hline
    \end{tabular}
    \end{center}
    \vspace{5pt}
\caption{\textbf{Ablation study on color video.} The table presents the PSNR/SSIM values of each model variant when trained on DAVIS dataset corrupted with Gaussian noise of level $\sigma$ = 30. The UDVD was trained using MSE loss as the noise model was unknown. The text in bold shows the model variant with the highest PSNR values. }
\label{tab:ablation}
\end{table}

\section{Discussion}

In this study, we proposed an innovative unsupervised video denoising framework that leverages a two-stage optimization strategy and a novel refine-net that employs the SIREN architecture as its backbone. Our model has been demonstrated to be highly effective across a range of video datasets, from natural videos to challenging fluorescence microscopy and calcium imaging recordings.

One key strength of our approach is its adaptability to various noise types and levels without needing prior knowledge about the noise characteristics. This flexibility is primarily due to our unsupervised training strategy which enables the model to learn directly from the noisy frames and adapt to the specific noise present in the video.

Moreover, the use of SIREN in our refine-net is another significant advantage, as it excels in representing complex patterns and structures in the data, making it particularly suitable for refining the denoised output.

However, like all models, ours also has certain limitations. The denoising quality could be affected by the value of $L$ in the feature generator. Setting a lower $L$ might result in blurry output frames, whereas a higher $L$ could potentially lead to overfitting. Therefore, the selection of $L$ requires careful tuning based on the specific characteristics of the dataset.

Another challenge arises when dealing with extremely noisy video sequences. Our model might struggle with recovering high-quality frames from these videos as the noise might overwhelm the actual content in the frames. Future research could focus on developing more robust denoising models that can handle higher noise levels.

In conclusion, our proposed video denoising framework presents a promising approach to tackle the video denoising problem, and we hope that our work will inspire future research in this direction.

\small
\bibliographystyle{plain}
\bibliography{denoise}

\newpage
\appendix
\appendix
\section{Network Architecture and Training}

\subsection{Feature Generator}
\begin{table}[h]
    \begin{center}
    \begin{tabular}{ccc}
     \hline
     \centering
      Name &  $N_{out}$  & Function\\ 
      \hline
      Input &  $2\times C_{coord} \times L$  & $\gamma(p_t)$  \\ 
      conv\_0 &  256  & Convolution 1 x 1 \\
      conv\_1 &  256  & Convolution 3 x 3 \\
      conv\_2 &  256  & Convolution 3 x 3 \\
      conv\_3 &  256  & Convolution 3 x 3 \\
      conv\_4 &  256  & Convolution 3 x 3 \\
      conv\_5 &  $C_{feat}$  & Convolution 3 x 3 \\
     \hline
    \end{tabular}
    \end{center}
\caption{\textbf{Network architecture of the feature generator}}
\label{tab:feat}
\end{table}

The feature generator $\mathcal{F}_\theta$ takes in a batch of sampled coordinated grids $\{G_t | t = 1,...,B\}$ corresponding to a batch of noisy frames $\{I_t | t = 1, 2, \dots, B\}$, where $B$ is the batch size.
In our experiment, we set $B = 5$, and we used 2 consecutive frame coordinate grids before and after the central frame, resulting in 5 frame coordinate grids including the central frame itself. Our goal is to generate a batch of feature map, denoted as $\{F_t | t = 1,...,B\}$, corresponding to each of the input frame.

Initially, a positional encoding $\gamma(.)$ is applied to these batched coordinate grids before they are processed by the feature generator. $\gamma(.)$ expands the low-dimensional coordinates into a high-dimension feature space using the following equation:
\begin{equation}
\gamma(\mathbf{p}_t) = [\mathrm{sin}(2^0 \pi \mathbf{p}_t), \mathrm{cos}(2^0 \pi \mathbf{p}_t), ..., \mathrm{sin}(2^{L-1} \pi \mathbf{p}_t), \mathrm{cos}(2^{L-1} \pi \mathbf{p}_t)]
\label{pos}
\end{equation}
where $L$ signifies the frequency level of output feature map. The output size of $\gamma(.)$ depends on the size of input coordinate $C_{coord}$ and $L$.
Next, the feature generator $\mathcal{F}_\theta$ takes positional encoded grid $\gamma(G_t)$ as input, and output the feature map $F_t\in\mathbb{R}^{H \times W \times C_{feat}}$.

\subsection{Denoisier}

\begin{table}[h]
    \begin{center}
    \begin{tabular}{ccc}
     \hline
     \centering
      Name &  $N_{out}$  & Function\\ 
      \hline
      Input &  $B*C_{feat}$  &   \\ 
      conv\_0 &  256  & Convolution 1 x 1 \\
      conv\_1 &  96  & Convolution 1 x 1 \\
      conv\_2 &  $C_{in}$  & Convolution 1 x 1 \\
     \hline
    \end{tabular}
    \end{center}
\caption{\textbf{Network architecture of the Denoiser $\mathcal{D}_\phi$}}
\label{tab:denoise}
\end{table}

The network architecture for the Denoiser $\mathcal{D}_\phi$ is described in Table \ref{tab:denoise}. 
The batched feature map output from $\mathcal{F}_\theta$ is concatenated to form an input dimension of $1 \times B*C_{feat} \times H \times W$, and then fed into $\mathcal{D}_\phi$. The output of the Denoiser network is $C_{in}$, where $C_{in} =3$ for color images and $C_{in}=1$ for grayscale images. 


\subsection{Refine-Net}

\begin{table}[h]
    \begin{center}
    \begin{tabular}{ccc}
     \hline
     \centering
      Name &  $H_{out}$  & Function\\ 
      \hline
      Input &  $C_{coord}$ &   \\ 
      mlp\_0 &  256  & Linear \\
      mlp\_1 &  256  & Linear \\
      mlp\_2 &  256  & Linear \\
      mlp\_3 &  256  & Linear \\
      mlp\_out &  $C_{in}$  & Linear \\
     \hline
    \end{tabular}
    \end{center}
\caption{\textbf{Network architecture of the Refine-Net.}}
\label{tab:ablation}
\end{table}

The network architecture of the Refine-Net $\mathcal{F}_\theta$ is depicted in Table \ref{tab:ablation}. The backbone of this network is a Sinusoidal Representation Networks (SIREN) as described in \cite{siren}. This network takes as its input the coordinates of the denoised output from the denoise-net and produces a refined denoised central frame where each pixel coordinate is denoted by $\textbf{p}_t = (x,y,t)$.

The Refine-Net is composed of fully connected layers, which include 4 hidden layers, with the activation function at each layer being a sine function. Each hidden layer contains 256 units, and the output layer is composed of $C_{in}$ nodes, where $C_{in}=3$ for RGB images and $C_{in}=1$ for grayscale images. The refined denoised output is generated by processing the denoised image's coordinates through this network, resulting in a more precise final image.


\subsection{Training Details}

Our proposed network is trained using a two-stage process to effectively leverage the functionalities of each individual component - the feature generator, denoiser, and refiner. In the first stage, we train the feature generator and denoiser together as a combined model. Here, we optimize for the reduction of noise in the video frames. The aim at this stage is to extract relevant spatio-temporal features from the noisy inputs and then utilize those features to output an initial, denoised estimate of the central frame of the batch.

In the second stage of training, we keep the weights of the feature generator and denoiser fixed, and focus on training the refiner. The refiner uses the denoised output from the previous stage and enhances it further, particularly focusing on the high-frequency details. 

This two-stage training approach ensures an effective noise reduction while preserving and enhancing the structural details in the video sequences. It is important to note that each stage of training involves its own data augmentation strategy and the whole network is trained end-to-end, with the weights of each component optimized for their specific tasks. By doing so, our network provides an overall pipeline that delivers high-quality denoised video frames.

\textbf{Optimization Details:} 
The training of our network leverages the Adam optimizer \cite{kingma2014adam}. Initially, the feature generator and denoise-net are jointly trained, minimizing $l_1$ loss between the central channel of the feature map and corresponding noisy frames, along with the $l_1$ loss between the denoised central frame and its corresponding original frame. The balance between these two loss terms is controlled by $\lambda_1$ in \ref{loss}, which is set to 1.0.
\begin{equation}
    \mathcal{L}_{1} = ||\hat{I}_B - I^c||_1 + \lambda_1(\frac{1}{B}\sum_{t=1}^B|| F_t^{c} - I_t)||_1
    \label{loss}
\end{equation}

Subsequently, we refine the denoised output using the refine-net, minimizing the $l_1$ loss between the refine-net's output and both the denoise-net's output and the corresponding noisy frame. The regularization terms $\lambda_2$ and $\lambda_3$, set to 0.1 and 1.0 respectively, help to strike a balance between detail preservation and noise reduction.
\begin{equation}
L_{2} = \lambda_2||\mathcal{R}_\eta(G_t^c) - I^c||_1 + \lambda_3||\mathcal{R}_\eta(G_t^c) - \hat{I}_B||_1
\label{refine_loss}
\end{equation}

The initial learning rates are set to $0.0001$ for the joint training stage and $0.00001$ for the refine-net training. Both rates are reduced by a factor of 10 after every 1000 epochs. During the refine-net training stage, the weights of the feature generator and denoise-net are frozen.


\section{More Visual Comparison of Denoising Algorithms}

In this section, we present a comprehensive visual comparison of our denoising method against the UDVD \cite{udvd}, utilizing several test datasets. Our experiments use the DAVIS and Set8 datasets, each corrupted with various noise types: Gaussian noise with $\sigma = 30$ and $50$, Poisson noise with $\lambda=30$ and $50$, and Impulse noise with $\alpha = 0.2$. The denoising performance is evaluated and visually demonstrated on several distinct scenarios, including 'bus', 'hypersmooth', and 'snowboard'. 

\begin{figure}[h]
  \centering
  \includegraphics[width=0.9\textwidth]{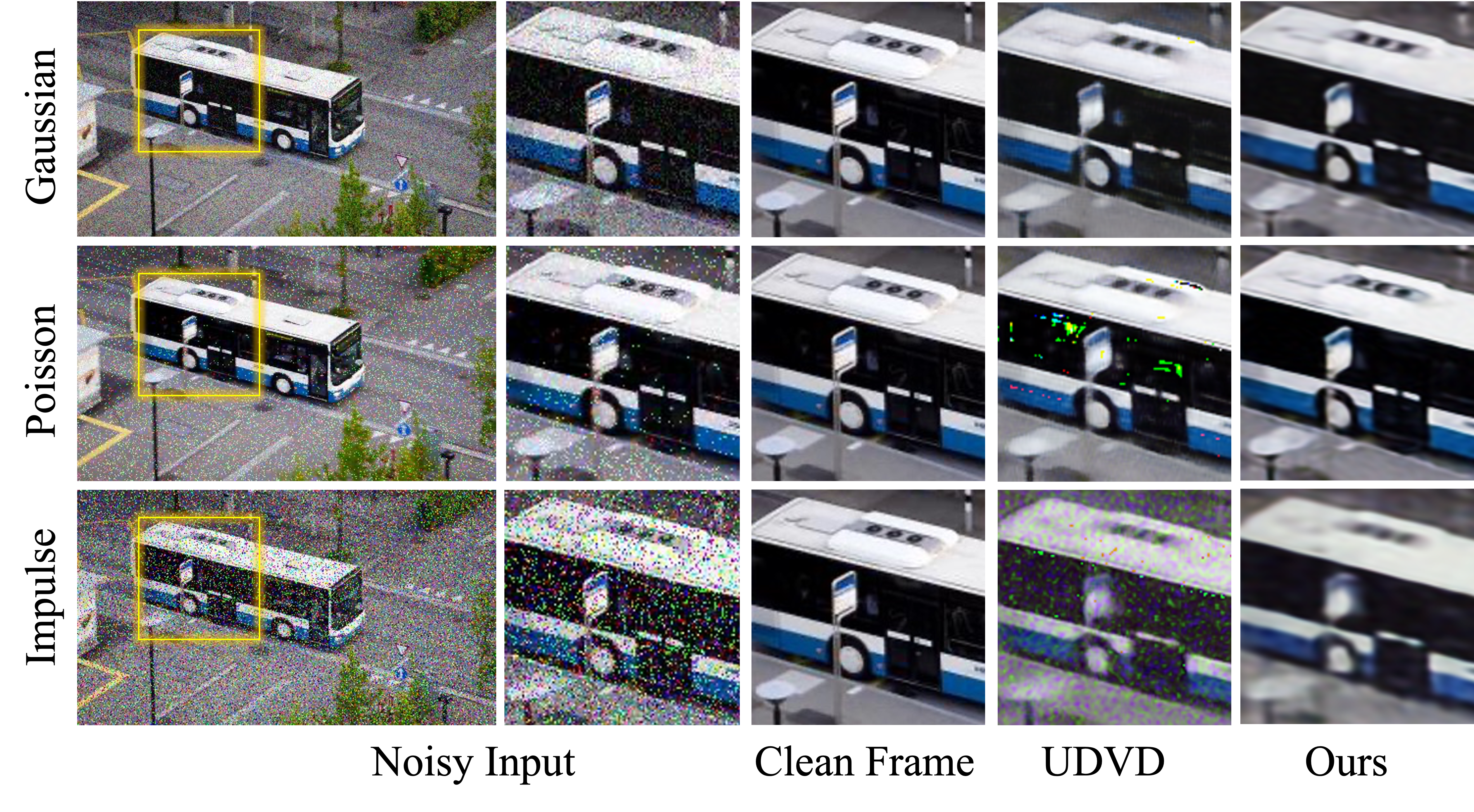}
  \label{fig:supp_bus}
  \caption{Visual comparison on the denoising result of bus video sequence in the DAVIS dataset}
\end{figure}

\begin{figure}[h]
  \centering
  \includegraphics[width=0.9\textwidth]{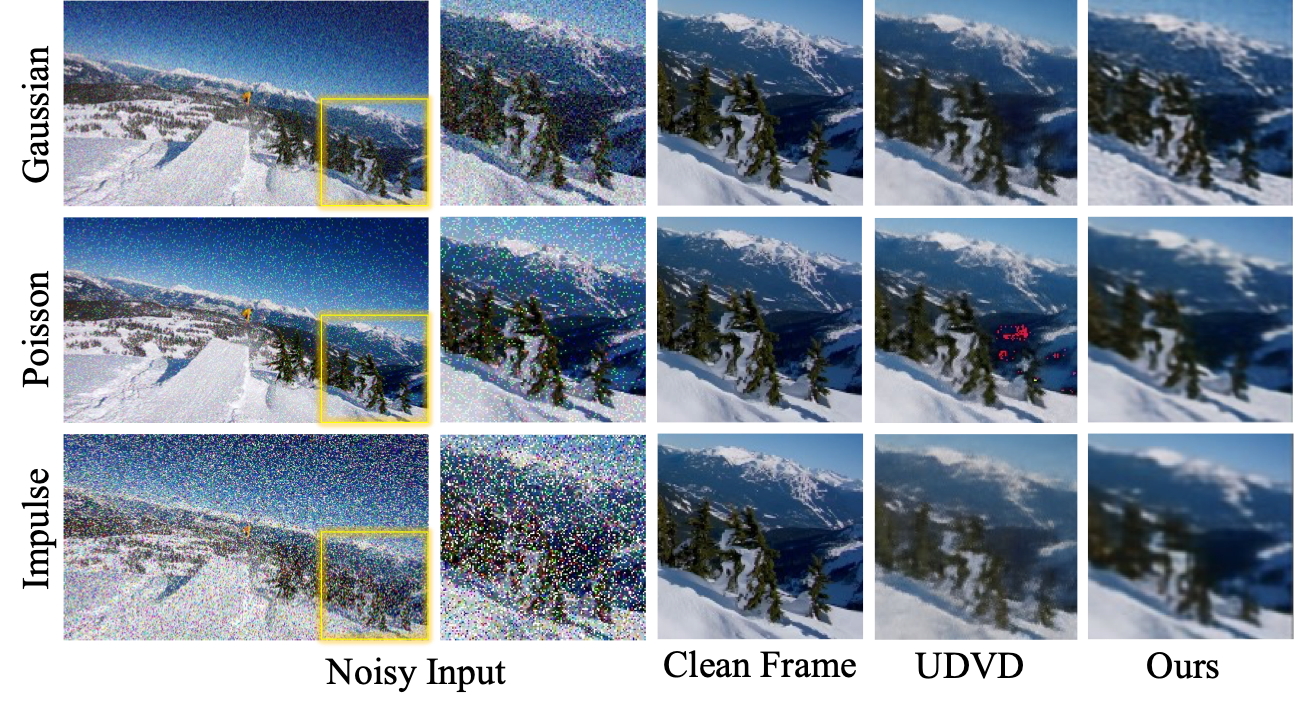}
  \label{fig:supp_bus}
  \caption{Visual comparison of the denoising result of snowboard video sequence in SET8 dataset}
\end{figure}

\begin{figure}[h]
  \centering
  \includegraphics[width=0.9\textwidth]{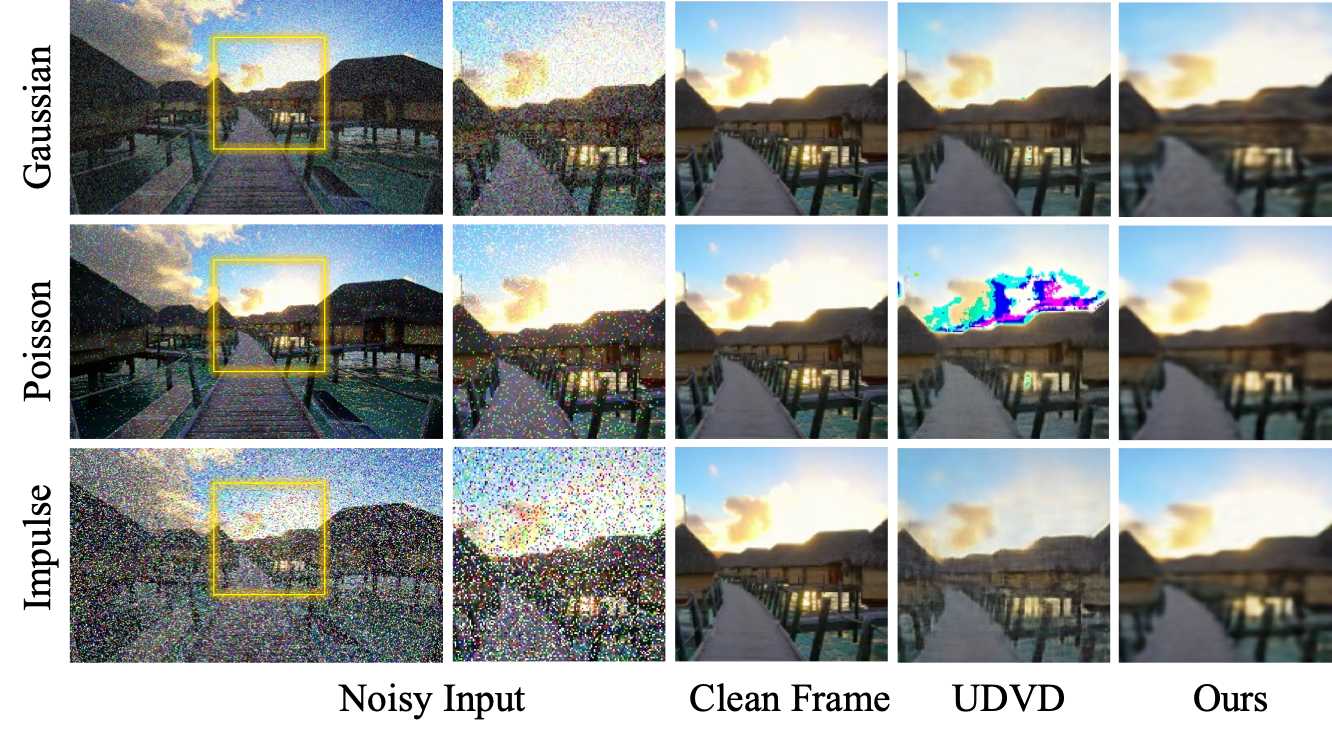}
  \label{fig:supp_bus}
  \caption{Visual comparison of the denoising result of hypersmooth video sequence in SET8 dataset}
\end{figure}

\section{Ablation Study}
\textbf{Positional encoding.} We provide ablation studies on the positional encoding hyperparameters in \ref{pos} using video sequences in DAVIS dataset. We experimented with different frequency levels in \ref{pos} to encode the input coordinates. Table \ref{tab:PE} describes the average PSNR and SSIM value using different $L$:
\begin{table}[h]
    \begin{center}
    \begin{tabular}{cccccccc}
     \hline
     \centering
       &Metrics &  $L=5$  &  $L=10$ &  $L=20$ &  $L=30$ &  $L=50$ &  $L=100$\\ 
      \hline
     \multirow{2}{*}{\parbox{1.2cm}{Poisson $\lambda=30$}} & PSNR &  27.84 &  27.54 &  27.46 & \textbf{29.01}  & 27.46  & 27.30  \\ 
      & SSIM & 0.78   & 0.77 &  0.77 &  \textbf{0.81} & 0.77  & 0.76  \\
     \hline
     \multirow{2}{*}{\parbox{1.2cm}{Gaussian $\sigma=30$}} & PSNR &  \textbf{27.18} &  26.18 &  26.12 & 26.74  & 25.81  & 26.04  \\ 
      & SSIM & 0.74   & 0.72 &  0.72 &  \textbf{0.75} & 0.71  & 0.72  \\
     \hline
    \end{tabular}
    \end{center}
\caption{\textbf{Postional encoding ablation} $L$ means frequency level used to encode the low dimensional coordinate to high dimension. The texts in bold indicate the highest value.}
\label{tab:PE}
\end{table}


\end{document}